\documentclass[10pt, a4paper]{article}
\usepackage{lrec}
\usepackage{graphicx}
\usepackage{tabularx}
\usepackage{soul}

\usepackage{epstopdf}
\usepackage[latin1]{inputenc}

\usepackage{hyperref}
\usepackage{xstring}

\usepackage{comment, algorithm, amsmath, amsfonts, amsthm}
\usepackage{graphicx}
\usepackage{algpseudocode}
\algrenewcommand\algorithmicrequire{\textbf{Input:}}
\algrenewcommand\algorithmicensure{\textbf{Output:}}

\usepackage{color}
\usepackage{multirow}

\theoremstyle{definition}

\newtheoremstyle{myprob}%
{}{}{\normalfont}{}{\scshape}{.\,}{ }{}
\theoremstyle{myprob}
\newtheorem{problem}{Problem}

\newcommand{\citet}{\newcite}
\newcommand{\citep}{\cite}

\title{Unsupervised Domain Adaptation of Language Models \\ for Reading Comprehension}

\name{Kosuke Nishida, Kyosuke Nishida, Itsumi Saito, Hisako Asano, Junji Tomita}

\address{NTT Media Intelligence Laboratories, NTT Corporation\\
		1-1 Hikarinooka Yokosuka-Shi, Kanagawa, Japan\\
         kosuke.nishida.ap@hco.ntt.co.jp\\}

\abstract{
This study tackles unsupervised domain adaptation of reading comprehension (UDARC). Reading comprehension (RC) is a task to learn the capability for question answering with textual sources. State-of-the-art models on RC still do not have general linguistic intelligence; i.e., their accuracy worsens for out-domain datasets that are not used in the training. We hypothesize that this discrepancy is caused by a lack of the language modeling (LM) capability for the out-domain. The UDARC task allows models to use supervised RC training data in the source domain and only unlabeled passages in the target domain. To solve the UDARC problem, we provide two domain adaptation models. The first one learns the out-domain LM and in-domain RC task sequentially. The second one is the proposed model that uses a multi-task learning approach of LM and RC. The models can retain both the RC capability acquired from the supervised data in the source domain and the LM capability from the unlabeled data in the target domain. We evaluated the models on UDARC with five datasets in different domains. The models outperformed the model without domain adaptation. In particular, the proposed model yielded an improvement of 4.3/4.2 points in EM/F1 in an unseen biomedical domain. \\ \newline  }
\begin{document}

\maketitleabstract

	\section{Introduction}
    Reading comprehension (RC) is a task to acquire a capability of understanding natural language for question answering with textual sources.
    It has seen significant progress since the release of numerous
    datasets such as SQuAD \citep{squad} and the rise of the deep neural models such as BiDAF \cite{bidaf}. 
    Recently, fine-tuning of pre-trained language models (LM) such as BERT \cite{bert} has achieved state-of-the-art performance in many NLP tasks including RC.
    
    However, such state-of-the-art models still do not have general linguistic intelligence; e.g., their accuracy is sensitively affected by the difference in the distribution between the training and evaluation datasets, such as in the domains of textual sources.
    This discrepancy becomes an issue in a real-world scenario.
    For instance, a business intending to introduce an RC application for a service must create tens of thousands of (passage, query, answer) tuples per domain of the service \cite{linguistic}.
    However, such a large annotation to create training data costs much money and takes up a lot of time.
	When the domain entails personal information or expert knowledge, even crowd-sourcing can not be used to create the training data.
    
    This study tackles a task, called \textbf{Unsupervised Domain Adaptation of Reading Comprehension} (UDARC). 
    Table \ref{goal} shows the task setting of UDARC.
    The model can use the (passage, query, answer) tuples of the source domain for training. In addition, the model can use unlabeled passages for training in the target domain. 
    Annotated corpora with QA pairs in the target domain can not be used for training. 
    In addition to the interest in the domain discrepancy, we think that this setting is a natural one in real-world scenarios, where it is assumed that the RC application provider has documents that can be used as knowledge sources for RC in the target domain (e.g., technical documents about a product).
    UDARC thus enables people who have such documents in the target domain but have no RC training data for the domain to introduce RC applications. This scenario also applies to low-resource languages, where the availability of RC training data is limited.

	\begin{table}[t]
	\begin{center}
		\scalebox{0.8}{
			\begin{tabular}{r|ccc|ccc}\hline
				& \multicolumn{3}{c}{Training} & \multicolumn{3}{c}{Evaluation}\\
				& \multicolumn{2}{c}{Input} & Output 
				& \multicolumn{2}{c}{Input} & Output \\
                & Passage & Query & Answer & Passage & Query & Answer \\ \hline
            Source & \checkmark & \checkmark & \checkmark & & & \\ 
            Target & \checkmark & & & \checkmark & \checkmark & \checkmark \\ \hline
		\end{tabular}}
	\end{center}
	\caption{Task setting of unsupervised domain adaptation for reading comprehension (UDARC). Only unlabeled passages can be used for training in the target domain. In terms of QA pairs, the task requires zero-shot domain adaptation.
    }
	\label{goal}
	\end{table}

    We hypothesize that the poor performance of in-domain models for out-domains is caused by a lack of LM capability for out-domains. That is, we can improve the question answering accuracy, without the RC data for the out-domains, by training the language model with textual sources about the out-domains in an unsupervised fashion. 
    In this study, we introduce a no-adaptation baseline model. 
    It transfers a BERT model fine-tuned with the source domain RC dataset to the target domain without domain adaptation.
    As a natural unsupervised domain adaptation approach, we investigate a sequential model. 
    It adapts the language model of BERT with the unlabeled passages in the target domain, and then it fine-tunes BERT with the source domain RC dataset.

    Moreover, we propose the multi-task learning approach of LM in the target domain and RC in the source domain. It is more promising because the model avoids forgetting about the target domain while fine-tuning.
    This study investigates the feasibility of UDARC and the effectiveness of these models on various domain datasets.
	Our main contributions are as follows.
	\begin{itemize}
		\item We tackled
		unsupervised domain adaptation of reading comprehension (UDARC).
		To solve UDARC, we hypothesize that the ability to 
		understand the out-domain passages can be learned without supervised annotation for RC.
	    \item 
        We propose a multi-task learning approach of the LM in the target domain and the RC in the source domain. This approach retains both knowledge of RC in the source domain and knowledge of LM in the target domain without forgetting. 
        We use BERT as the backbone model.
		\item We evaluated 
		three models on UDARC with
		five datasets in different domains.
        The domain adaptation models outperformed the no-adaptation model. In particular, the proposed model (adapted from the Wikipedia domain 
		to a biomedical domain)
		yielded the best performance, with a 4.3/4.2 point gain in EM/F1 over the model without domain adaptation.
	    \item We thoroughly investigated cases in which UDARC 
	    is effective. 
	    We found that the task is 
	    promising when the target domain is unseen in the training of the source domain and the pre-training of the language model and when 
	    the training data in the target domain are insufficient to acquire the RC and LM capabilities.
	\end{itemize}

	\section{Problem Formulation}
    \label{sec:problem}
    Here, we will focus on unsupervised domain adaptation of extractive RC, which is the most popular problem formulation of RC. The UDARC task is defined as follows.
    \begin{problem}
    \label{prob:prob}
    Let $I_S$ be instances of 3-tuples (passage, query, answer span) in the \textit{source} domain, and $I_T$ be unlabeled passages in the \textit{target} domain.
    The task of UDARC is 
    to accurately answer the query (extract the correct answer-span from an input passage) in the target domain by training an extractive RC model with $I_S$ and $I_T$.
    \end{problem}

    \section{Related Work}
	\subsection{Reading Comprehension}
    \citet{synnet} and \citet{adamrc} also tackled UDARC. Their approaches are to create pseudo QA pairs for training in the target domain. 
    They use the answer extraction model to find a potential answer from the passage and the query generation model to create the query given the potential answer and the passage.
    Our approach is different from theirs because, according to our hypothesis, we can acquire knowledge in the target domain without RC training.
    An advantage of our approach is the low computational cost due to the lack of a need for the training of the Sequence-to-Sequence model to generate the query.
	
	The MRQA 2019 shared task has a similar motivation as ours \citep{mrqa}. Its goal is to generalize to new test distributions and be robust to test-time perturbations. The task is on six in-domain training data and twelve out-domain evaluation data (a many-to-many setting). Our motivation is to acquire the ability to understand out-domain passages from unlabeled passages and the ability to answer a query from annotated RC training data for domain adaptation (a one-to-one setting). This motivation corresponds to a real-world scenario. The shared task resulted in the certification of the dependence of RC performance on the backbone LM capability. This result supports the effectiveness of our approach to improve LM for domain adaptation.

    RC is essentially classified into four types: extractive \cite{squad}, multiple-choice \cite{race}, generative \cite{marco}, and cloze-style \cite{cnn}. 
    Although this study focused on extractive RC, we believe that the concept of our model can be applied to other types of RC.
    
    Many RC datasets have been published recently, but the ones for closed domains are limited in number. There are datasets, for example, in biomedical domain \cite{clicr}, scientific domain \cite{arc,sciq}, and software domain \cite{quasar}.
	The limited number is one reason for our developing UDARC. Despite the demand for RC in closed domains, which requires expert knowledge, the annotation cost of the dataset makes it difficult to create the dataset.

	\subsection{Domain Adaptation}
	\begin{figure*}[t!]
		\begin{center}
		\begin{tabular}{c}
            
            \begin{minipage}{0.15\hsize}
            \begin{center}
                \includegraphics[clip, height=4cm]{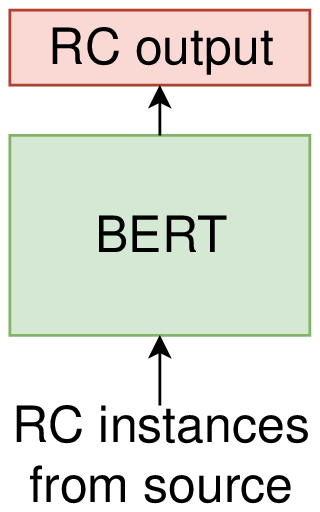}\\
                (a) No-adaptation
            \end{center}
            \end{minipage}
            
            \begin{minipage}{0.4\hsize}
            \begin{center}
                \includegraphics[clip, height=4cm]{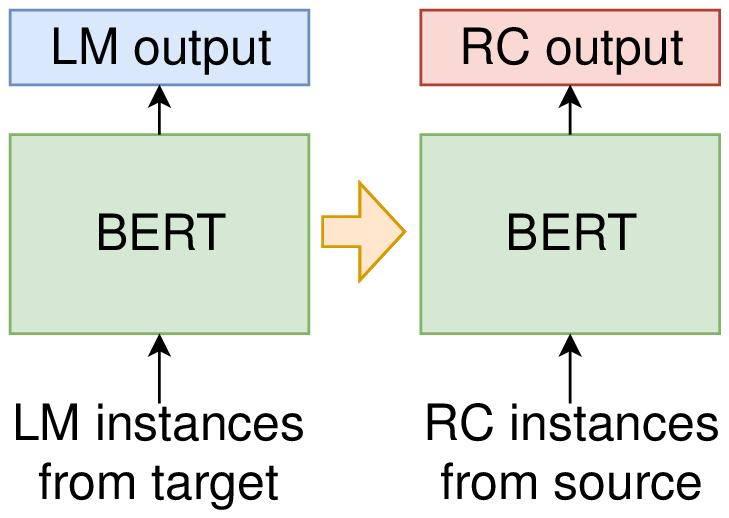}\\
                (b) Sequential
            \end{center}
            \end{minipage}

            \begin{minipage}{0.45\hsize}
            \begin{center}
                \includegraphics[clip, height=4cm]{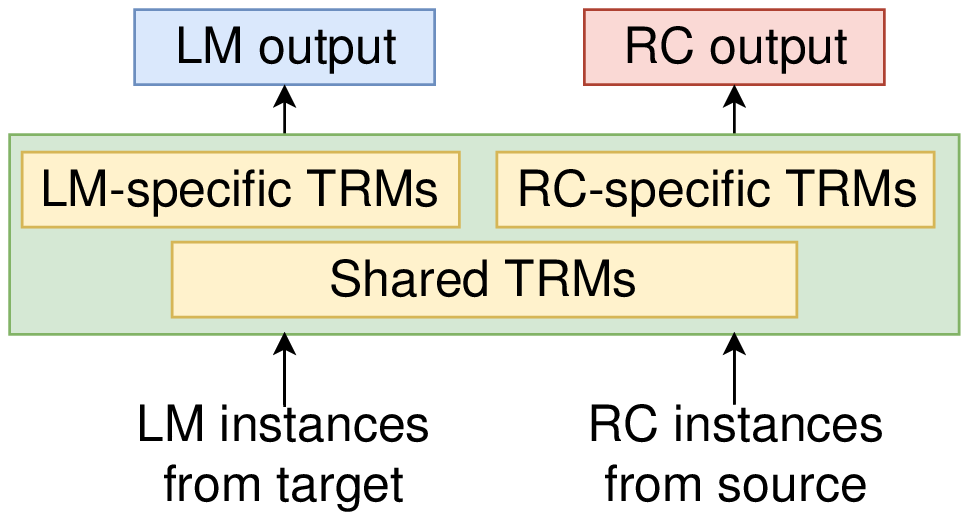}\\
                (c) Multi-Task
            \end{center}
            \end{minipage}
		\end{tabular}
		\caption{Overview of the three models. The LM instances are from the target domain, and the RC instances are from the source domain. ``TRMs" means transformer layers.}
		\label{model}
		\end{center}	
	\end{figure*}
	
	Unsupervised domain adaptation is a task to adapt the model to the target domain with labeled source data and unlabeled target data. While supervised domain adaptation with labeled target data first trains the model in the source domain and then adapts it to the target domain, unsupervised domain adaptation takes another approach. In NLP, \citet{UDA_NLP1} uses a two-step algorithm; the model first learns the representations and then learns the classification. \citet{UDA_NLP2} uses the multi-task learning approach of the classification and the feature selection of structural correspondence learning \citep{scl}. Here, \citet{MSDA_DAN} trains the representation learning and the classification jointly with a domain-adversarial neural network.

    However, UDARC can not use standard unsupervised domain adaptation methods. As shown in Table \ref{goal}, an instance is divided into an input (e.g., a (passage, query) pair in RC) 
    and output (e.g., an answer in RC). 
    Unsupervised domain adaptation allows inputs to be used in the target domain without outputs. In RC, neither the output nor the queries in the inputs can be used for training.

    Zero-shot learning is a task to adapt the model so that it can predict unseen classes in the training \citep{zsl1}. In particular, the recently proposed zero-shot domain adaptation \citep{zsda1,zsdda} can not use the inputs for the training of domain adaptation. UDARC can be interpreted as a kind of zero-shot domain adaptation, because it can not use queries in the inputs for the training.
	
	Zero-shot domain adaptation is a more challenging task than unsupervised domain adaptation, and there are few studies on it. \citet{zsdda} allows task-irrelevant data in the target domain. This is similar to our setting, but their task is classification of images, which is rather different from extractive RC, which is a special case of sequence labeling for text.
	The results of other studies have limitations when they are used for UDARC.
    \citet{zsda1} and \citet{zsda2} hypothesize that the training data are from multiple domains. \citet{zsda3} suppose there is prior knowledge about what factors cause the differences between the source and target data distributions.
	
	\section{Methods}
	This section introduces three models: the baseline model without domain adaptation, the sequential domain adaptation model, and the proposed model using multi-task learning.
	We used a pre-trained BERT$_\textrm{base}$ model as the backbone architecture for the three models. It was trained with large-scale corpora: BookCorpus (800M words) \citep{book} and English Wikipedia (2,500M words). It has achieved state-of-the-art performance on various RC datasets. See \cite{bert} for the details of BERT$_\textrm{base}$.
	Note that the models can use other pre-trained LM consisting of stacked layers as their backbone architecture.

    \subsection{No-adaptation Baseline}
	This baseline model is a BERT fine-tuned with an RC dataset in the source domain.
	It does not consider domain adaptation, so its performance corresponds to the lower bound of UDARC.

    
    Figure \ref{model} (a) shows an overview of the model.
    We follow the fine-tuning setup for extractive RC, as in \cite{bert}.
    We add a linear layer for extractive RC on top of the BERT layers, where its output dimension is two.
    The first dimension represents a score that the token is the start of the answer span, and the other dimension represents the end.
    The input sequence is [`[CLS]'; \textit{query}; `[SEP]'; \textit{passage}; `[SEP]'], where `[CLS]' and `[SEP]' are special tokens and `;' means concatenation. 
    
    
    \subsection{Sequential Model}
    This model first adapts the pre-trained BERT with unsupervised passages in the target domain. Then, it fine-tunes the adapted BERT with the RC dataset of the source domain.  
    
    Figure \ref{model} (b) shows an overview of the model.
    For the first unsupervised adaptation, a linear layer for LM is added at top of the BERT layers.  The BERT including the LM output layer is trained (with the sentences in the unsupervised passages) in the same manner as in the pre-training of BERT, i.e., by using masked language modeling (MLM) and next sentence prediction (NSP). 
    The input sequence is [`[CLS]'; \textit{sentences} 1; `[SEP]'; \textit{sentences 2}; `[SEP]'].
    After the adaptation, another linear layer for RC is added, and the BERT including the RC output layer is fine-tuned in the same manner as in the no-adaptation baseline.
    
    Note that we may conduct domain adaptation in reverse order, i.e., first build an RC model in the source domain and then adapt it to the target domain. However, this order causes catastrophic forgetting of the capability of finding an answer span. The sequential model follows the pipeline strategy of the previous unsupervised domain adaptation work \citep{UDA_NLP1}. Their model learns representations first and classification after that, but their task is not RC and they do not use a pre-trained language model.
	
    \subsection{Proposed Multi-Task Model}
    We consider that the sequential model forgets the knowledge in the target domain while it is being fine-tuned with RC training data.
    Moreover, it is known that multi-task fine-tuning of the pre-trained language models outperforms single-task fine-tuning \cite{mt-dnn}.
	For this reason, we propose a multi-task learning approach.
	Multi-task learning adds an RC linear layer and LM linear layer to the top of the BERT layers.
    The model uses the RC linear layer
	for RC instances and the LM linear layer for LM instances as the output layer.
	Figure \ref{model} (c) shows an overview of the model.
	
	Our multi-task learning approach uses the following two techniques.
	
	\subsubsection{Using Shared and Specific Transformer Layers}
	First, we use the top-$n$ Transformer layers (TRMs) for the tasks separately as task-specific RC or LM layers. The task-specific layers are cloned and initialized from the original pre-trained BERT$_\textrm{base}$ layers.
	The other layers are commonly shared by the tasks.
	The RC instances pass through the shared layers, the RC-specific TRMs, and the RC output layer.
	The LM instances pass through the shared layers, the LM-specific TRMs, and the LM output layer.
	This idea follows \citet{BERTlayers}'s observation that basic syntactic information (e.g., part-of-speech tagging) is captured in lower layers and  high-level semantic
    information (e.g., coreference labeling) are captured in higher layers.
    We consider that the basic syntactic information is the common features between tasks, and the high-level semantic information is closely related to the output of the task.
    Therefore, the LM instances in the target domain should use the shared TRMs in order to capture the basic syntactic information in the target domain.
    The LM instances should not share the higher layers of BERT with the RC instances, because the way it uses the high-level semantic information is different from in RC.
	
	\subsubsection{One-Segment LM Training}
	Second, we preprocess each LM instance as a sequence with one segment.
	That is, each LM instance is [`[CLS]'; `[LM]'; \textit{passage}; `[SEP]'], and the segment ids are a zero vector.
	Therefore, NSP is not used for training of our multi-task model.
	This intends to prevent the learning of the segment interaction in NSP from disturbing the learning of query-passage interaction, because the two segments only interact in the RC training.
	`[LM]' is a special token meaning that the instance is an LM instance.
	
	\subsubsection{Training Procedure}
	We perform the RC training and LM training alternately.
	The RC training is performed $k$ times as many times as the LM training.
	The algorithm is shown in Algorithm \ref{alg}.
	
	\begin{algorithm}[t!]
		\caption{Multi-task learning approach}
		\label{alg}
		\begin{algorithmic}[1]
			\Require source RC instances $I_S$, target LM instances $I_T$, 
			 num. of steps $N$, the RC training ratio $k$
			\ForAll{$i$ in $1, \cdots , N$}
			\State Select mini-batch $b_S \sim I_S$
			\State Train the shared layers, RC-specific layers, and RC output layer with $b_S$.
			\If{$i \% k ==0$}
			\State Select mini-batch $b_T \sim I_T$
			\State Train the shared layers, LM-specific layers, and LM output layer with $b_T$.
			\EndIf
			\EndFor
		\end{algorithmic}
	\end{algorithm}
    
    Note that the model size and computational time in the evaluation are the same as in the original BERT, because the LM-specific layers are not used for evaluation.
    In the training, the computational time remains about the same, because the training for each mini-batch is the same as in the original BERT. 
	
    \section{Experiments}
    \subsection{Dataset}
    We evaluated the UDARC task on various datasets.
    The training data in the source domain should be large and cover a wide range of topics.  Moreover, the test data in the target domain should be from a closed domain. Here, we selected five datasets from different domains. Table \ref{data} shows the statistics. We used the development data for the evaluation because some of the datasets do not include test data. Note that not all the unlabeled passages were used. The number of used unlabeled passages depended on the experimental setup (the number of fine-tuning epochs and the value of $k$) and the size of the training dataset.
    \begin{table}[t]
	\begin{center}
			\scalebox{0.95}{
			\begin{tabular}{ccccc}\hline
            dataset & domain & \# training & \# dev. & \# unlabeled \\
             & & data & data & passages \\ \hline
            SQuAD & Wikipedia & 87599 & 10570 & 19047\\
            NewsQA & news & 107064 & 5988 & 95933\\
            BioASQ & biomedical & 0 & 1504 & 55148 \\
            DuoRC & movie & 69524 & 15591 & 5137 \\
            Natural & HTML & \multirow{2}{*}{104071}& \multirow{2}{*}{12836} & \multirow{2}{*}{12222}\\
            Questions & Wikipedia &  & & \\ \hline
            \end{tabular}}
	\end{center}
	\caption{Statistics of the datasets used in the experiments.}
	\label{data}
	\end{table}
    \begin{table*}
    \begin{center}
        \begin{tabular}[t]{l|cc|cccc} \hline
         & Train w./ & Domain & \multicolumn{4}{c}{Target} \\
         & SQuAD & Adaptation & NewsQA & BioASQ & DuoRC & NQ\\ \hline
        Standard RC in Target & & & 41.5/56.0 & --- & 20.2/27.2 & 58.9/72.2 \\ \hline
            No-adaptation & \checkmark & & 35.2/50.7 & 41.1/53.6 & 24.5/33.0 & \textbf{44.4}/\textbf{57.5} \\
        Sequential & \checkmark & \checkmark & 35.2/51.0 & 44.5/57.1 & 25.4/33.8 & --- \\
        Multi-Task & \checkmark & \checkmark & \textbf{35.9}/\textbf{51.4} & \textbf{45.4}/\textbf{57.8} & \textbf{25.5}/\textbf{34.1} & 43.8/56.7 \\ \hline
        \end{tabular}
    \caption{Results when the SQuAD was the source dataset. EM is on the left and F1 is on the right in each cell. The top row is the supervised training in the target domain, so it is the expected upper bound of UDARC. The Standard RC in BioASQ is empty, due to the lack of training data.
    The sequential model for NQ is empty. NSP of the BERT pre-training can not be applied to HTML with a lot of HTML tags (e.g., List and Table).
    }
    \label{from_squad}
    \end{center}
    \end{table*}
    \begin{table*}
    \begin{center}
        \begin{tabular}[t]{l|cc|cccc} \hline
        & Train w./ & Domain & \multicolumn{4}{c}{Target} \\
        & NewsQA & Adaptation & SQuAD & BioASQ & DuoRC & NQ \\ \hline
        Standard RC in Target & & & 80.9/88.4 & --- & 20.2/27.2 & 58.9/72.2 \\ \hline
        No-adaptation & \checkmark & &  59.8/73.9 & 34.5/48.3 & 22.5/31.2 & 39.0/52.7 \\
        Sequential & \checkmark & \checkmark & 59.7/75.3 & 36.6/\textbf{50.4} & 23.7/\textbf{32.7} & --- \\
        Multi-Task & \checkmark & \checkmark & \textbf{60.6}/\textbf{75.8} & \textbf{36.8}/50.3 & \textbf{23.8}/32.3 & \textbf{42.0}/\textbf{56.2} \\ \hline
        \end{tabular}
    \caption{Results when the NewsQA was the source dataset. 
    }
    \label{from_news}
    \end{center}
    \end{table*}
	
	\vspace{.5em}\noindent\textbf{SQuAD1.1} is  an RC dataset from Wikipedia \cite{squad}. We used this dataset as the source domain or target domain.
	We used the passages of the training data as the unlabeled passages when the dataset was in the target domain.
	
	\vspace{.5em}\noindent\textbf{NewsQA} is an RC dataset from CNN news \citep{newsqa}.
	We used this dataset as the source domain or target domain.
	We used the CNN news scripts \citep{cnn} as unlabeled passages similar to the data collection of NewsQA.
	
	\vspace{.5em}\noindent\textbf{BioASQ} is 
	a biomedical semantic indexing and question answering challenge\footnote[1]{Task 7b, Biomedical Semantic QA, is held with ECML PKDD 2019. See http://BioASQ.org/ .} \citep{bioasq}.
    The MRQA 2019 shared task preprocesses this dataset for extractive RC; we used the MRQA version of this dataset. 
    The training data of this dataset is not provided in the MRQA 2019 shared task.
    We used this dataset only for the target domain.
    We collected the unlabeled passages from the abstracts of PubMed articles.  
    
    Our main interest in the experiments is the performance of the unsupervised domain adaptation models
    on BioASQ. This is because the BioASQ domain is not fully covered by the source domain (Wikipedia or news) or by the BERT pre-training (BookCorpus and Wikipedia).
    
    
    \vspace{.5em}\noindent\textbf{DuoRC} is an RC dataset in the movie domain \citep{duorc}. DuoRC provides parallel movie plots from Wikipedia and IMDb. In the experiment, we used the ParaphraseRC task in DuoRC, where each query is created on a Wikipedia movie article, and each passage is collected from an IMDb article corresponding to the same movie.
    The ParaphraseRC task is difficult because it is designed to contain a large number of queries with low lexical overlap between queries and their corresponding passages.
    We used this dataset only for the target domain and the passages of the training data as the unlabeled passages. 
    
    The movie domain is different from the source domains, whereas the BERT pre-training covers many stories in BookCorpus. Therefore, we think that the pre-trained BERT has acquired knowledge of language modeling for this dataset.
    

    
    \vspace{.5em}\noindent\textbf{Natural Questions (NQ)} is an RC dataset containing passages from Wikipedia written in HTML format, where each passage is given as a sequence of words and HTML tags \citep{nq}. 
    We used the preprocessed dataset for extractive RC in the MRQA 2019 shared task for the training and evaluation.
    We used the passages of the training data in the original NQ as the unlabeled passages.
    
    Although the domain is the same as that of SQuAD and is also covered by the BERT pre-training, we used this dataset to confirm whether our model can adapt to HTML format without supervised RC data on the HTML format.

    \subsection{Experimental Setup}
    We compared three models, no-adaptation, sequential, and multi-task on the above datasets.
    
	We used the PyTorch implementation of BERT\footnote[2]{ https://github.com/huggingface/pytorch-transformers}.
	We trained the models on four NVIDIA Tesla P100 GPUs. 
	The optimizer was Adam \cite{adam}. 
	The warm-up proportion was 0.1 and the learning rate was 0.00005. The batch size was 32.
	There were three epochs.
	The input length was 384. Sequences longer than the input length were truncated with a stride length of 128. The other hyperparameters followed those of BERT$_\textrm{base}$.
	The ratio of RC training to LM training $k$ is 10.
	The number of task-specific layers $n$ is 3.
	The hyperparameters are fixed in all of the fine-tunings for RC.
	The hyperparameters of the pre-training in the sequential model follow the default settings of the implementation, except that the input length is 512, which is the longest case.
	
    We evaluated the answer prediction in terms of exact match (EM) and partial match (F1). These are official metrics of SQuAD.
    
    \subsection{Results}
    \paragraph{Under what condition is UDARC effective?}
    Table \ref{from_squad} (the source domain is Wikipedia from SQuAD) and Table \ref{from_news} (news from NewsQA) show the performance of the models for the target domain.
    The ``Standard RC in Target" row lists 
    the results in the standard RC (non-UDARC) setting, where each model was trained and evaluated in the target domain, so these are expected to be the upper bounds.
    First, we discuss each target-only dataset separately.
    
    \vspace{.5em}\noindent\textbf{BioASQ.} BioASQ is of main interest in our experiments. 
    The two domain adaptation models outperformed the no-adaptation baseline in both the source domain settings. The proposed model improved on the no-adaptation model trained in SQuAD by 4.3/4.2 points.

    
    BioASQ is in a domain that is not included in the BERT pre-training corpora. The results showed that the UDARC framework is effective in adapting to unseen domains.
    Moreover, this result confirms that our hypothesis behind UDARC is correct; i.e., the ability to understand the out-domain passages can be learned from a non-annotated corpus and the ability to answer the query can be learned from annotated RC training data, even though BERT is pre-trained with very large corpora.
    
    \vspace{.5em}\noindent\textbf{DuoRC.}
    All unsupervised domain adaptation models outperformed the model trained with the supervised dataset in the target domain\footnote[3]{The performance of the supervised model was 20.2/27.2, which is similar to the performance (19.7/27.6) reported in the original paper.}. 
    This surprising result is caused by the difficulty of the ParaphraseRC task of DuoRC. It has low lexical overlap between queries and their corresponding passages, and 
    the passages are rather long and complicated. 
    As a result, the training data are too difficult for learning the RC capability.
    We think UDARC is promising 
     when it is difficult to acquire the LM and RC capabilities with the supervised datasets.

    \vspace{.5em}\noindent\textbf{Natural Questions.}
    We compared four models (No-adaptation / Multi-Task with the source domain of SQuAD / NewsQA) in the target domain of NQ. The results indicated that unsupervised adaptation from SQuAD to NQ did not improve accuracy; on the other hand, unsupervised adaptation from NewsQA to NQ was effective. These results can be interpreted as meaning that the multi-task approach trained in the news domain as the source successfully adapted to the Wikipedia domain. However, the proposed model trained with plain text failed to adapt to HTML format. We consider that the adaptation to the HTML format is a more challenging task than domain adaptation.
    Here, the task design of the language modeling remains as future work to understand text in HTML format, such as understanding of the dependencies among segments separated by HTML tags.
    
    \paragraph{What is the performance of the three models?}
    Here, let us discuss the model performances shown in Table \ref{from_squad} and Table \ref{from_news}. Except for NQ, the domain adaptation models outperformed the no-adaptation baseline.
    In terms of the EM metric, the multi-task model outperformed the sequential model for all source/target settings. This tendency showed the possibility that the sequential model forgets the out-domain knowledge while it is being fine-tuned.
    However, there was no statistical significance between the two models.
    We observed that the sequential model was as effective as the multi-task model. It is worth pre-training BERT with the unlabeled target domain
    data after pre-training in the general domain.
    This finding can be applied to other NLP tasks.
    
    In comparison to the related work, \citet{adamrc} only refers to the experiments with BERT in the setting from SQuAD to NewsQA. The improvement of their domain adaptation method over BERT fine-tuned in the source domain is 0.6/0.5 points, which is comparable to our 0.7/0.7 point improvement, though their pseudo QA generation approach requires more computational cost.
    
    \paragraph{Does domain adaption hurt the performance in the source domain?}
    \begin{table}[t]
        \begin{center}
            \scalebox{0.9}{
            \begin{tabular}[t]{lcccc} \hline
             & NewsQA & BioASQ & DuoRC & NQ \\ \hline
            Standard RC & \multicolumn{4}{c}{80.9/88.4} \\ \hline
            Sequential & 81.2/88.6 & 80.6/88.4 & 81.0/88.4 & --- \\ 
            Multi-Task & 81.1/88.5 & 81.1/88.5 & 80.7/88.3 & 80.9/88.4 \\ \hline
            \end{tabular}}
        \end{center}
        \caption{Results for when the source and evaluation datasets were SQuAD. The model was adapted to each target dataset. The top row is the no-adaptation model trained only with the SQuAD dataset. }
        \label{squad_after}
    \end{table}
    \begin{table}[t]
        \begin{center}
            \scalebox{0.9}{
            \begin{tabular}[t]{lcccc} \hline
            & SQuAD & BioASQ & DuoRC & NQ \\ \hline
            Standard RC & \multicolumn{4}{c}{41.5/56.0} \\ \hline
            Sequential & 41.8/56.9 & 42.0/57.1 & 42.7/58.0 & --- \\
            Multi-Task & 42.6/57.6 & 42.0/57.0 & 42.8/57.8 & 42.3/57.4 \\ \hline
            \end{tabular}}
        \end{center}
        \caption{Results for when the source and the evaluation datasets were NewsQA. The model was adapted to each target dataset. The top row is the no-adaptation model trained only with the NewsQA dataset. 
        }
        \label{news_after}
    \end{table}
    
    We evaluated 
    the drop in performance in the source domain due to the domain adaptation.
    Table \ref{squad_after} and Table \ref{news_after} show the results.
    The ``Standard RC" row is the model trained only in the source domain.
    
    Surprisingly, the sequential 
    model and the multi-task model tend to outperform the standard RC training in the source domain.
    In addition to the discrepancy between domains, a discrepancy also exists between the training samples and the evaluation samples.
    We consider that the domain adaptation has the effect of generalization, so the model overcomes the sampling discrepancy.
    UDARC requires no additional supervised data for training, so the framework of UDARC can be easily extended to the standard RC task in which we can expect an improvement in performance.
    
\begin{figure}[t!]
		\begin{center}
        \includegraphics[clip, width=7.5cm]{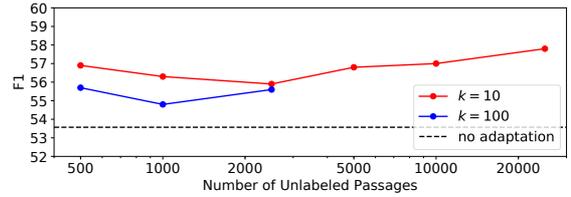}\\
		\caption{Performance (F1) of the proposed model in adaptation from SQuAD to BioASQ in terms of the number of unlabeled passages. The proposed model was trained with $k$ times fewer LM instances than RC instances. The right-most coordinate of each line is the maximum number of used LM instances determined by $k$, the number of epochs, and the number of training data.}
		\label{num_doc}
		\end{center}	
	\end{figure}

    \paragraph{How many unlabeled passages are required?}
    We evaluated the performance of the proposed model in terms of the number of unlabeled passages. The experiments were on adaptation from SQuAD to BioASQ, because this setting showed the largest improvements and BioASQ is the main focus of UDARC. Figure \ref{num_doc} shows the results.
    
	The results show that the proposed model outperformed the no-adaptation baseline even when there were only 500 unlabeled passages. The gain was 3.7/3.3 points in EM/F1. In terms of the RC-LM ratio $k$, $k=10$ was preferred, but we should note that in some cases, $k=100$ was preferred in the pilot experiments. The biomedical domain is far from the pre-training corpus of BERT, so we consider that the moderate frequency of LM training steps is larger than in other domains included in the corpora.
	We found that the performance does not always increase as more unlabeled passages come to be used, though the best performance is with the full 26280 passages.
     
\begin{figure}[t!]
		\begin{center}
        \includegraphics[clip, width=7.5cm]{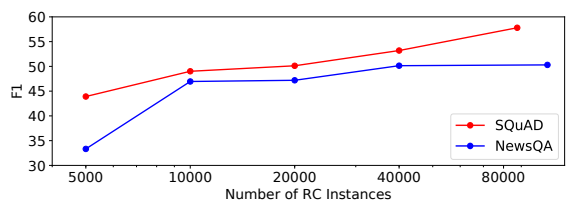}\\
		\caption{Performance (F1) of the proposed model in adaptation from SQuAD and NewsQA to BioASQ versus number of source training data.}
		\label{num_source_bio}
		\end{center}	
	\end{figure}

\begin{figure}[t!]
		\begin{center}
        \includegraphics[clip, width=7.5cm]{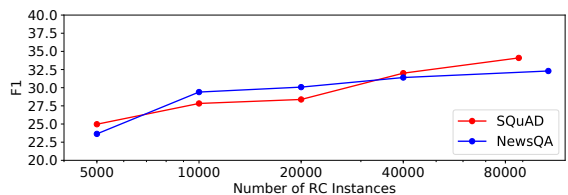}\\
		\caption{Performance (F1) of the proposed model in adaptation from SQuAD and NewsQA to DuoRC versus number of source training data.}
		\label{num_source_duo}
		\end{center}	
	\end{figure}

    \paragraph{What domain is preferred as the source domain?}
	To evaluate the preference about the source domain under the same conditions, we equalized the number of training data in the source domain.
    Figure \ref{num_source_bio} and Figure \ref{num_source_duo} shows the performance of the proposed model with BioASQ and DuoRC as the target dataset.
	
	On BioASQ, the proposed model adapted from SQuAD outperformed the model adapted from NewsQA. In contrast, on DuoRC, the performance was on par. 
    Therefore, the performance of the proposed model depends on the selection of the source domain, but the preferred source domain cannot as yet be identified.
	
    \paragraph{How much source data are required?}
	Figure \ref{num_source_bio} and Figure \ref{num_source_duo} show the performance in terms of the number of source training data.
    The results indicate that more training data results in higher performance.
    In particular, the improvement grows rapidly until 10000 instances, after which it becomes slower.
    This result coincides with the observation of \citet{linguistic}.
    They showed that tens of thousands of training data are required to fine-tune BERT.
    We consider that the required number of source data for UDARC shows the same tendency as in the standard RC task.
	
	\section{Conclusion and Future Work}
	This paper studied UDARC to adapt an RC model to the target domain without any annotated data in the target domain.
    
    This is the first study to focus on the unsupervised domain adaptation for acquiring the ability to answer the question from the RC task in the source domain and the ability to understand the out-domain passages from the LM task in the target domain.
    This approach is different from the related work, which generates the pseudo QA pairs for the training. We described two unsupervised domain adaptation models using BERT.
    In addition, the proposed model reduces the forgetting of out-domain knowledge while it is being fine-tuned.
    
    We evaluated the two models and the no-adaptation model on the five datasets in different domains. 
    As a result, the domain adaptation models outperformed the no-adaptation model especially well when the target domain was not contained in the source domain or the BERT pre-training corpora.
    The proposed model (adapted from the Wikipedia domain to the biomedical domain) yielded the best performance, with a 4.3/4.2 points gain in EM/F1 over the no-adaptation model.
    We believe that this study sheds light on the importance and feasibility of UDARC.
    Our experiments also showed that the UDARC framework has the potential to outperform a model trained with supervised datasets in the target domain
    when it is difficult to acquire the LM and RC capabilities from the supervised datasets.
    
    Pre-trained language models \cite{bert,gpt2,xlnet,roberta} and the fine-tuned models achieved state-of-the-art performance in many NLP tasks, including RC. We believe that this study considering unsupervised domain adaptation with BERT will foster great contributions to various fields of NLP that have not been the subject of previous work in unsupervised domain adaptation.

    \section{Bibliographical References}
	\bibliography{theme}
	\bibliographystyle{lrec}
\end{document}